\documentclass[10pt,journal,compsoc]{IEEEtran}

\usepackage{cite}
\usepackage{graphicx}
\usepackage{amsmath}
\usepackage{amssymb}
\usepackage{booktabs}
\usepackage{multirow}

\usepackage[capitalize]{cleveref}
\crefname{section}{Sec.}{Secs.}
\Crefname{section}{Section}{Sections}
\Crefname{table}{Table}{Tables}
\crefname{table}{Tab.}{Tabs.}
\newcommand{\etal}{\textit{et al}.}

\newcommand{\eg}{\textit{e}.\textit{g}.}

\begin{document}

\title{APRNet: Attention-based Pixel-wise Rendering Network for Photo-Realistic Text Image Generation}

\author{Yangming Shi \thanks{Yangming Shi is with the Automation Department of University of Science and Technology of China. (E-mail: {ymshi}@mail.ustc.edu.com)}, Haisong Ding, Kai Chen, and Qiang Huo
\thanks{Haisong Ding, Kai Chen and Qiang Huo are with the Speech group of Microsoft Research Asia, Beijing, China (E-mail: \{hadin, kaic, qianghuo\}@microsoft.com).}
\thanks{
This work was done when Yangming Shi worked as the intern in Speech Group, Microsoft Research Asia, Beijing, China.}}


\IEEEtitleabstractindextext{%
\begin{abstract}
Style-guided text image generation tries to synthesize text image by imitating reference image's appearance while keeping text content unaltered. The text image appearance includes many aspects. In this paper, we focus on transferring style image's background and foreground color patterns to the content image to generate photo-realistic text image. To achieve this goal, we propose
   1) a content-style cross attention based pixel sampling approach to roughly mimicking the style text image's background;
   2) a pixel-wise style modulation technique to transfer varying color patterns of the style image to the content image spatial-adaptively; 
   3) a cross attention based multi-scale style fusion approach to solving text foreground misalignment issue between style and content images;
   4) an image patch shuffling strategy to create style, content and ground truth image tuples for training.
    Experimental results on Chinese handwriting text image synthesis with SCUT-HCCDoc and CASIA-OLHWDB datasets demonstrate that the proposed method can improve the quality of synthetic text images and make them more photo-realistic.
\end{abstract}

}

\maketitle

\IEEEdisplaynontitleabstractindextext
\IEEEpeerreviewmaketitle

\IEEEraisesectionheading{\section{Introduction}\label{sec:introduction}}

\IEEEPARstart{S}{tyle-guided} text image generation is a challenging task, which tries to synthesize text images by imitating reference style image's appearance while keeping text content unaltered (\eg \cite{SCGAN_2020_ICFHR, textstylebrush_2021_arxiv, kang_2021_TPAMI}).
The appearance of a text image includes foreground and background color patterns, spatial transformations and deformations, typography (for printed text), writing styles and stroke thickness (for handwriting text), etc \cite{ textstylebrush_2021_arxiv}. 
In this paper, we propose an \textbf{A}ttention-based \textbf{P}ixel-wise \textbf{R}endering \textbf{Net}work (APRNet) to transfer the foreground and background color patterns from a style text image to a new image without altering its text content.    

Previously, text image generation is mainly based on traditional signal processing techniques, which has been successfully applied to augment optical character recognition (OCR) model training in natural scene scenarios (\eg \cite{SynText-NIPS-2014, SynText-CVPR-2016, SynText-IJCV}).
These kinds of methods synthesize text images by rendering text transcriptions with given typeset fonts, followed by colorization, distortion, and natural scene image blending. 
The synthetic image style is controlled by pre-defined rules and not photo-realistic.

Recently, neural network based style-guided approaches have been applied to printed \cite{editing_2019_MM, Shimoda_2021_ICCV} and handwriting text image generation \cite{SCGAN_2020_ICFHR, textstylebrush_2021_arxiv, kang_2021_TPAMI}. Equipped by the generative adversarial network (GAN) \cite{GAN_2014_NIPS}, style transfer \cite{Gatys_2016_CVPR, perceptual_2016_ECCV, review_2019_TVCG, Adain_2017_CVPR, StyleGAN_2019_CVPR, StyleGAN2_2020_CVPR} and image-to-image translation \cite{pix2pix_2017_CVPR, cyclegan_2017_ICCV, unsupervised_2017_NIPS, MUNIT_2018_ECCV}, the quality of synthetic text images has been improved greatly. 
However, transferring varying foreground and background color patterns from camera-captured style images to content images at line-level is still an unsolved problem.
It is mainly caused by complicated backgrounds, various lighting conditions and foreground (text) misalignment. 

To improve the color pattern transfer quality of text image generation, especially at line-level, we make the following technical contributions:

\begin{itemize}
    \item [1)] An \textbf{Att}e\textbf{n}tion-based \textbf{Pix}el S\textbf{amp}ling (\textbf{AttnPixamp}) module is proposed to roughly mimicking style text image's background;
    \item [2)] A \textbf{Pix}el-wise St\textbf{y}le \textbf{Mod}ulation (\textbf{PixyMod}) approach is derived based on StyleGANv2\cite{StyleGAN2_2020_CVPR} to transfer spatial-varying color patterns;
    \item [3)] An \textbf{Att}e\textbf{n}tion-based \textbf{Mul}ti-scale \textbf{S}tyle \textbf{F}usion (\textbf{AttnMuSF}) module is designed based on spatial pyramid pooling \cite{SPP_2015_TPAMI} and cross attention to solve text foreground misalignment issue between style and content images;
    \item [4)] An image patch shuffling strategy named \textit{Single Crop} is used to train this well-designed APRNet in a self-supervised manner;
    \item [5)] Experimental results on Chinese handwriting text image synthesis with SCUT-HCCDoc \cite{HCCDoc_2020_PR} and CASIA-OLHWDB \cite{CASIA_2011_ICDAR} datasets demonstrate that the proposed method can improve the quality of synthetic text images and make them more photo-realistic.
\end{itemize}

The rest of this paper is organized as follows. In \cref{sec:related_work}, we review previous style transfer and text image generation works. In \cref{sec:method}, we introduce our APRNet design in detail. Experimental results are presented in \cref{sec:exp} and limitations of our method are discussed in \cref{sec:lim}. Finally we conclude the paper in \cref{sec:con}.

\section{Related Work}
\label{sec:related_work}
\subsection{Style transfer and exemplar-guided generation}
Style-guided handwritten text generation is a sub-field of style transfer, whose goal is to mimic the reference image's appearance and transfer expected styles into given content images. Gatys \etal \cite{Gatys_2016_CVPR} perform artistic style transfer by jointly minimizing the feature and style reconstruction loss (using Gram Matrix to represent styles). StyleGAN \cite{StyleGAN_2019_CVPR} employs AdaIN for style integration, while StyleGANv2 \cite{StyleGAN2_2020_CVPR} further improves generation quality by activation standard deviation (std) modulation and normalization. The latest version of StyleGANv3 \cite{StyleGAN3_2021_arxiv} presents a principled solution to aliasing caused by pointwise nonlinearities by considering their effects in a continuous domain. Besides, GuaGAN \cite{SPADE_2019_CVPR} proposes a spatially-adaptive normalization residual block to catch spatial-varying styles. 

Another similar task is exemplar-guided image synthesis, whose applications include pose-guided person generation, face attributes editing, virtual try-on, sketch image colorization, etc. The algorithms \cite{exemplar_2019_ICLR, SC-GAN_2021_ICCV, Adaattn_2021_ICCV, CoCosNet_2020_CVPR, CoCosNet2_2021_CVPR, DAGAN_2020_MM, selectiongan_2019_CVPR, AlBahar_2019_ICCV, Park_2019_CVPR} range from explicit image matching by giving additional conditions to implicit learning using flexible cross-attention models. However, unlike these methods which pay more attention to the global similarity between references and outputs, text image generation cares both the style realism of global and local regions and the rightness of content.

\subsection{Text image generation} 
\textbf{Handwritten text image generation.} Handwritten text generation can be traced to 2014, when the generative adversarial network \cite{GAN_2014_NIPS} was proposed to generate handwritten digital symbols from a noise vector. Afterwards, the cGAN \cite{cGAN_2014_arxiv} allows controlling the class of generated contents by inputting an additional class condition. The latest techniques \cite{higan_2021_AAAI, HWT_2021_ICCV, kang_2021_TPAMI, JokerGAN_2021_MM, Scrabblegan_2020_CVPR, ganwriting_2020_ECCV, SmartPatch_2021_ICDAR} focus on line-level style-guided text image generation. These systems synthesized glyph-realistic text line images with different writer styles, usually optimized by discriminators, writer classifiers, and character recognizers. Besides, the metaHTR \cite{MetaHTR_2021_ICCV} attempted to use a meta-learning framework to achieve writing style adaptive generation with few data. However, most previous works pay more attention to handling diverse glyphs while not considering complex style patterns in reality. 
In this regard, skeleton-based approaches avoid the glyph burden and focus more on style transfer. SC-GAN \cite{SCGAN_2020_ICFHR} employs the StyleGAN's decoder to integrate style code into content code and render photo-realistic text images in simple cases. Coincidentally, TextStyleBrush \cite{textstylebrush_2021_arxiv} distills text image content from its appearance, which can be applied to the new content with self-supervised ways. However, these methods still cannot handle complex foreground and background color patterns due to the limitation of employed style vectors.

\textbf{Printed text image generation.} Printed image generation remains a hot topic. Early studies \cite{SynText-NIPS-2014, SynText-CVPR-2016, SynText-IJCV} synthesized images by traditional signal processing techniques such as colorization, distortion, and natural scene image blending. However, these kinds of methods suffered none photo-realistic from pre-defined rules. The latest advances also benefit from GANs. Wu \etal \cite{editing_2019_MM} proposed an end-to-end style retention network (SRNet) with text conversion, background inpainting, and fusion for editing text in the wild. Shimoda \etal \cite{Shimoda_2021_ICCV} suggested a text vectorization technique by leveraging the advantage of differentiable text rendering to accurately reproduce the input raster text in a resolution-free parametric format. On the other hand, the printed character image generation \cite{MXFONT_2021_ICCV, DMFONT_2020_ECCV, LFFont_2021_AAAI, SMGAN_2021_TPAMI, Yang_2018_TIP, TE141K_2020_TPAMI, Yang_2019_IJCV} has been comprehensively studied. These methods simulate writing styles such as fonts, thickness, and artistic style while preserving the base structure. However, it is not recommended to directly employ related techniques for handwriting image generation task considering different application scenarios.

\section{Methodology}
\label{sec:method}

\subsection{Overview}
\begin{figure*}[htbp]
  \centering
  \includegraphics[width=\linewidth]{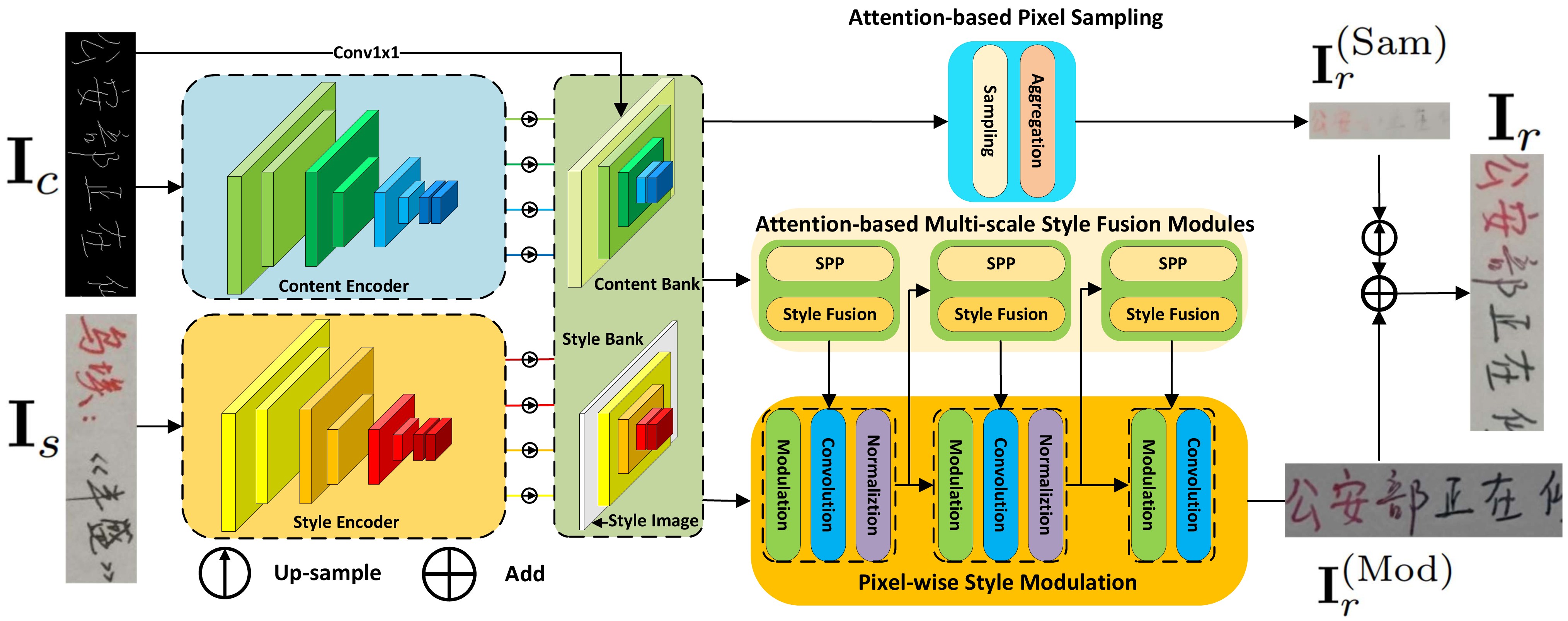}
  \caption{Overview of Attention-based Pixel Rendering Network. It consists of a style/content encoder to extract style/content encodings, an AttnPixamp module to render background, an AttnMuSF module to generate scale-fused style encodings, and a PixyMod module to render foreground and refine background spatial-adaptively. These designs will help to synthesize photo-realistic text images with spatial-varying color patterns.}
  \label{fig:overview}
\end{figure*}
In this paper, we explore solutions to render photo-realistic text images with style guidance for handwriting text image rendering. 
As shown in \cref{fig:overview}, our model takes a binary handwriting skeleton image (denoted as $\mathbf{I}_{c}$) as content input, and a real handwriting text line image (denoted as $\mathbf{I}_{s}$) as style reference input. The sizes of $\mathbf{I}_{c}$ and $\mathbf{I}_{s}$ are all $H\times W$.
In practice, the skeleton images can be rendered from online handwriting texts (\eg\cite{SCGAN_2020_ICFHR}), synthesized from generative models (\eg\cite{graves2013generating}), or extracted from real handwriting images (\eg\cite{Wang_Liu_2018}).  

We want to generate a rendered image $\mathbf{I}_{r}$, which shares the same foreground and background color patterns with $\mathbf{I}_{s}$, and its text content and character glyphs keep the same with $\mathbf{I}_{c}$. This is a challenging style transfer task due to complex backgrounds, colors, textures, noises, stroke thicknesses, and writing styles in $\mathbf{I}_s$. Besides, text foreground misalignment between $\mathbf{I}_s$ and $\mathbf{I}_c$ brings in more difficulties to achieve this goal.

During rendering, $\mathbf{I}_{c}$ and $\mathbf{I}_{s}$ are encoded first by a fully convolutional network (FCN) based content and style encoders, respectively.  
Extracted content and style feature maps are stored into content and style banks and serve as the inputs of the following 2-stage rendering procedure. 
The first rendering stage is called \textit{AttnPixamp}, which samples pixels from style images directly and aggregates them with a well-designed content-style cross attention mechanism to generate preliminary rendering results.
The second stage consists of two modules: an \textit{AttnMuSF} module, which fuses multi-scale style features to construct a better style encoding, 
and a \textit{PixyMod} module, which applies style modulation to content features spatial-adaptively to capture varying style patterns.
The outputs of these two stages will be fused by addition to synthesize a final rendered image $\mathbf{I}_{r}$.


\subsection{Content and style encoders}
\begin{figure}[htbp]
\centering
  \includegraphics[width=0.5\linewidth,angle=-90]{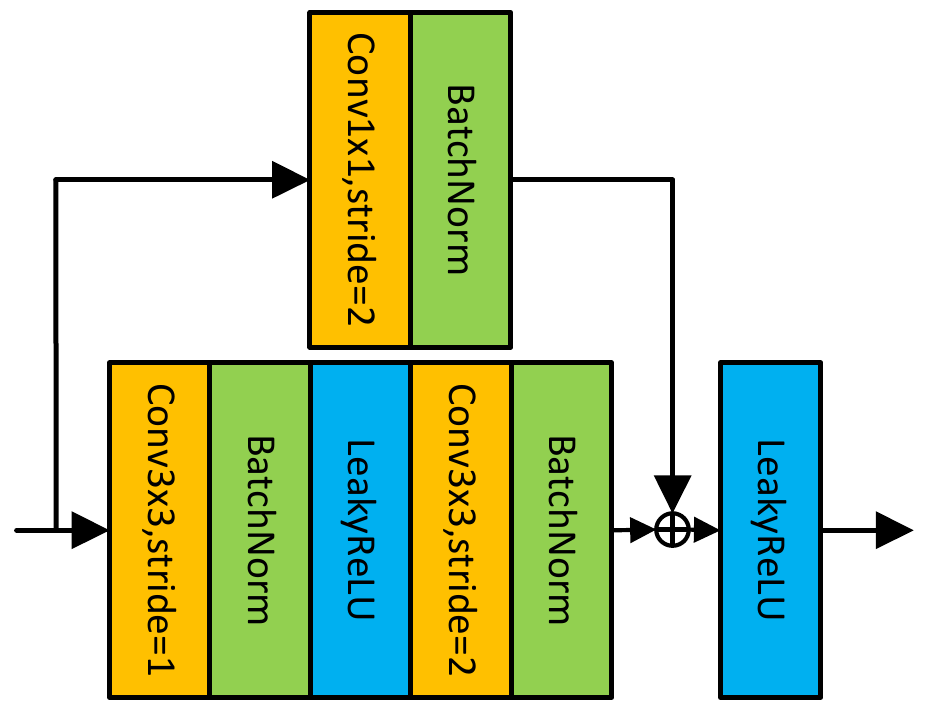}
  \caption{Blocks of one FCN encoder stage. Each encoder consists of 4 stages. All strides in convolution layers of the last stage are 1. For content encoder, the kernel size of highway convolution (the upper one) is $2 \times 2$.}
  \label{fig:encoder}
\end{figure}
The content and style encoders in our model share the same FCN architecture as shown in \cref{fig:encoder}. This architecture consists of 4 stages without using any max-pooling layers, and the output sizes of each stage are 
$\frac{H}{2}\times\frac{W}{2}\times32$,
$\frac{H}{4}\times\frac{W}{4}\times64$,
$\frac{H}{8}\times\frac{W}{8}\times128$ and
$\frac{H}{8}\times\frac{W}{8}\times256$, respectively. 
These output feature maps of the content/style encoder are stored in the aforementioned content/style bank as shown in \cref{fig:overview} for the next rendering procedures. 

Compared with SC-GAN \cite{SCGAN_2020_ICFHR}, which only uses the content encoder's last output as input to decoder and style encoder's last output to extract style vector, our rendering modules leverage early stage low-level features such as edge, texture, etc., and later stage high-level semantic features simultaneously. Low-level features are helpful for stroke generation, and high-level features are essential for background rendering.

In the content bank, we also store an $H\times W\times256$ tensor, which is extracted from $\mathbf{I}_c$ by a 256-channel $1\times1$ convolution (\cref{fig:overview}). This redundant operation is quite helpful to avoid stroke loss and get continuous strokes.

Since $\mathbf{I}_s$ is used in AttnPixamp module, it is also stored in the style bank (\cref{fig:overview}).

\subsection{Rendering stages}
\subsubsection{Attention-based pixel sampling}
\begin{figure}[htbp]
  \centering
  \includegraphics[width=\linewidth]{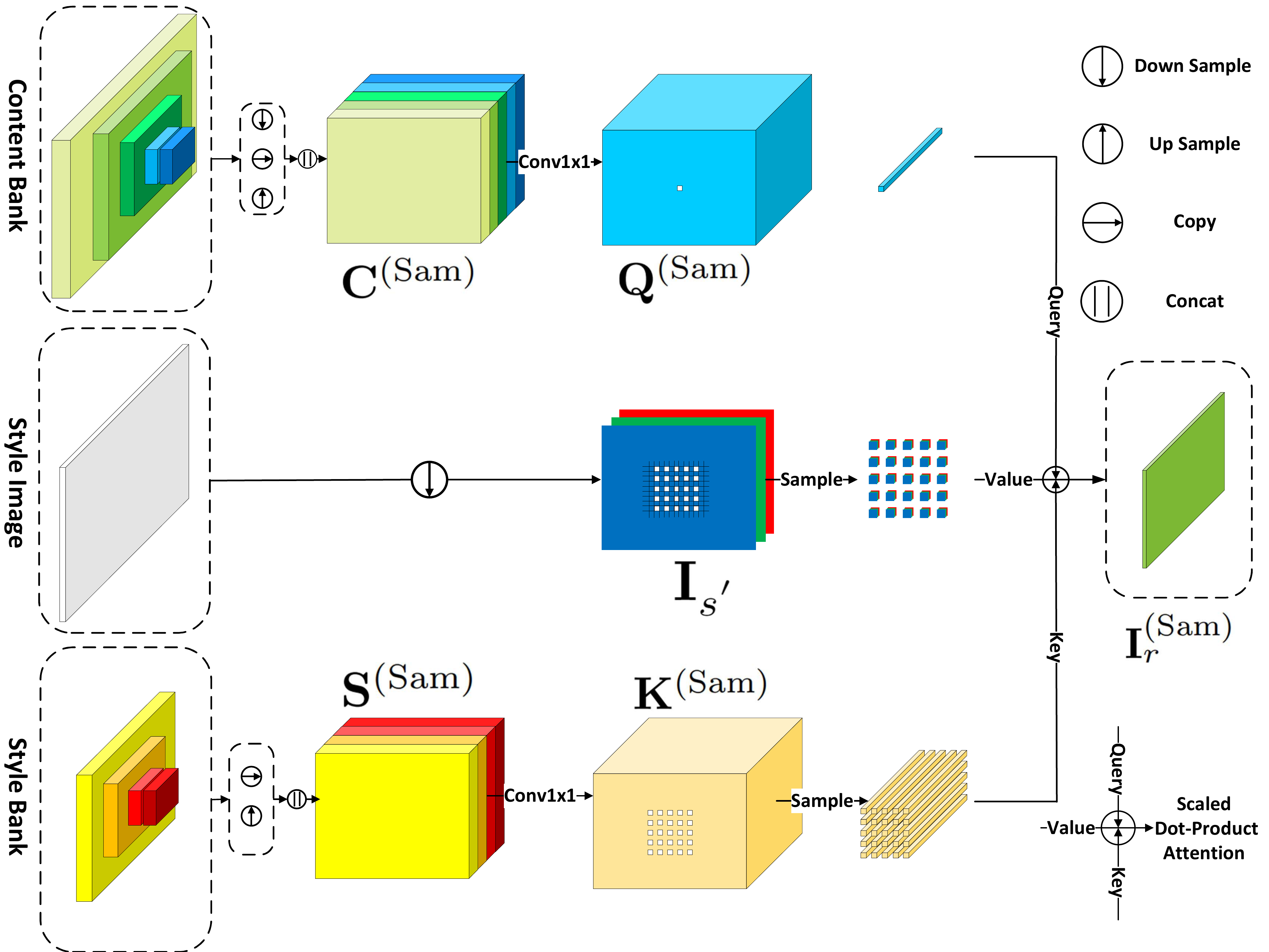}
  \caption{Architecture of attention-based pixel sampling module.}
  \label{fig:aps}
\end{figure}
Generally speaking, the majority of pixels in a text line image belong to background. Given a background pixel of $\mathbf{I}_c$, we can usually find background pixels from its respective neighborhood in $\mathbf{I}_s$. 
Based on this observation, it would be a more direct way to render $\mathbf{I}_{r}$'s background pixels by sampling from their respective neighborhoods in $\mathbf{I}_s$ compared with regressing RGB vectors through complex operations.

In practice, we first get a $\frac{H}{2}\times \frac{W}{2}$ style image $\mathbf{I}_{s^{'}}$ by down-sampling $\mathbf{I}_s$ to reduce following memory cost.
For coordinate $(i,j)$ in $\mathbf{I}_{s^{'}}$, $k^2$ pixels from its $(km+k-m)\times(km+k-m)$ square neighborhood in $\mathbf{I}_{s^{'}}$ are sampled uniformly, where $m$ is the horizontal/vertical distance between 2 successively sampled pixels (\cref{fig:aps}).
These sampled pixels are then aggregated by a content-style cross attention mechanism. We denote the rendered image in this step as $\mathbf{I}_{r}^{\mathrm{(Sam)}}$, whose size is also $\frac{H}{2}\times \frac{W}{2}$.

The attention mechanism is shown in \cref{fig:aps}, and it works as follows:
\begin{enumerate}
    \item [1)] Down-sample or up-sample all feature maps in content and style banks to  $\frac{H}{2}\times \frac{W}{2}$, and concatenate them respectively to get content tensor $\mathbf{C}^{\mathrm{(Sam)}}\in\mathbb{R}^{\frac{H}{2}\times \frac{W}{2}\times736}$ and style tensor $\mathbf{S}^{\mathrm{(Sam)}}\in\mathbb{R}^{\frac{H}{2}\times \frac{W}{2}\times480}$;
    
    \item [2)] Modify channel numbers of $\mathbf{C}^{\mathrm{(Sam)}}$ and $\mathbf{S}^{\mathrm{(Sam)}}$ to $d_s$ by $1\times1$ convolutions respectively, and the resulted tensors are denoted as $\mathbf{Q}^{\mathrm{(Sam)}}$ and $\mathbf{K}^{\mathrm{(Sam)}}$.
    
    \item[3)] Denote $k^2$ sampled style pixels as value tensor $\mathbf{V}_{i,j}^{\mathrm{(Sam)}}$, $k^2$ respective vectors in  $\mathbf{K}^{\mathrm{(Sam)}}$ as key tensor $\mathbf{K}_{i,j}^{\mathrm{(Sam)}}$, and the $(i,j)$-th vector in $\mathbf{Q}^{\mathrm{(Sam)}}$ as query vector $\mathbf{q}_{i,j}^{\mathrm{(Sam)}}$, we can get the rendered $\mathbf{I}_{r_{i,j}}^{\mathrm{(Sam)}}$ through scaled dot-product cross attention:
    \begin{equation}
        \mathbf{I}_{r_{i,j}}^{\mathrm{(Sam)}} = \left[\operatorname{\mathbf{Softmax}}(\frac{{\mathbf{q}_{i,j}^{\mathrm{(Sam)}}}^T\mathbf{K}_{i,j}^{\mathrm{(Sam)}}}{\sqrt{d_s}})\right]^T\mathbf{V}_{i,j}^{\mathrm{(Sam)}}
    \end{equation}.
    
\end{enumerate}
Repeat the above procedure for all coordinates to get $\mathbf{I}_{r}^{\mathrm{(Sam)}}$.


Since most sampled style image pixels are background ones, it is predictable that this module will focus on rendering the content image's background and not suitable for foreground stroke rendering.
We design other two modules to render foreground as well as fine-tune background pixels.

\subsubsection{Pixel-wise style modulation}
\begin{figure}[htbp]
\centering
  \includegraphics[width=0.8\linewidth]{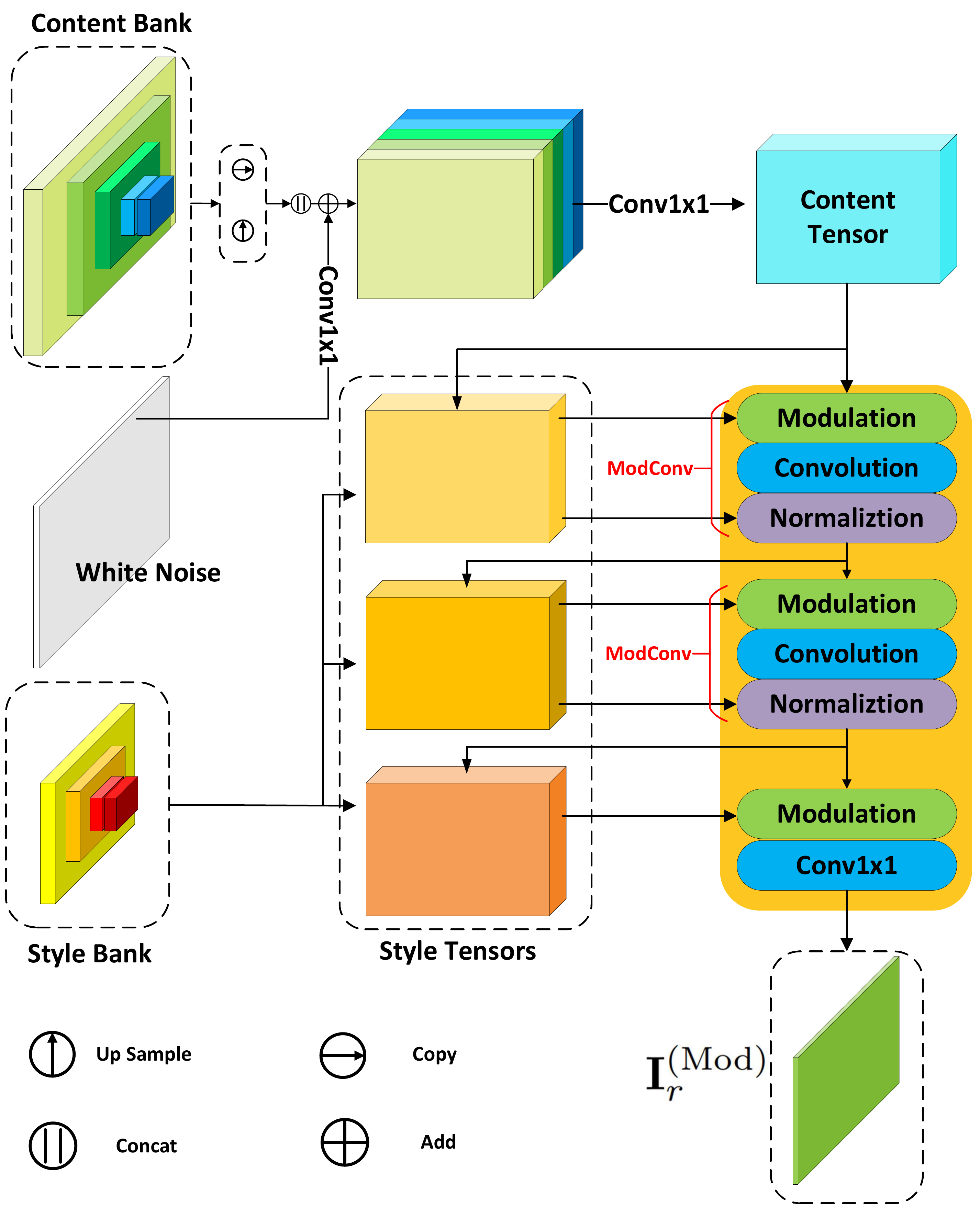}
  \caption{Details of pixel-wise style modulation. }
  \label{fig:psm}
\end{figure}
Previous works on text image style transfer usually encode a style input as a vector, and apply it to a generation procedure through AdaIN \cite{Adain_2017_CVPR}. 
It has been shown in \cite{StyleGAN2_2020_CVPR} that instance normalization may cause water droplet-like artifacts in generated images.
StyleGANv2 \cite{StyleGAN2_2020_CVPR} proposed to transfer styles through activation standard deviation (std) modulation and normalization to solve this problem and improve image quality.
However, it is still a style vector based approach, which means the transferred styles are uniform across spatial coordinates \cite{SPADE_2019_CVPR}. 
Experimental results in \cite{SCGAN_2020_ICFHR} and this paper also verify that for text image synthesis scenario, style vector based approach can only synthesize images with uniform color patterns.

It has been proven by SPADE\cite{SPADE_2019_CVPR} that applying style encoding spatial-adaptively is essential to catch spatial-varying styles.
Inspired by StyleGANv2 and SPADE, we propose a PixyMod module to transfer spatial-varying color patterns of the style image.

In StyleGANv2, the modulate and normalize activation std with style vector is equivalent to modulate and normalize weights of convolution layer, because the weights and style vector are all shared across spatial coordinates.
For PixyMod, we have to back-off to applying modulation and normalization on input and output tensors of convolution layer pixel-wisely.

Denote the content tensor as $\mathbf{C}\in\mathbb{R}^{H\times W\times I}$. Before input into convolution layer, it is first modulated by style tensor $\mathbf{S}\in\mathbb{R}^{H\times W\times I}$ element-wisely to get $\mathbf{M}\in\mathbb{R}^{H\times W \times I}$ as
\begin{equation}
    \mathbf{M} = \mathbf{S}\otimes\mathbf{C}
\end{equation}
where $\otimes$ is an element-wise multiplication.

Then, $\mathbf{M}$ is convoluted by $\mathbf{W}\in\mathbb{R}^{O\times k_h \times k_w\times I}$ to get $\mathbf{U}\in\mathbf{R}^{ H \times W \times O}$:
\begin{equation}
    \mathbf{U} = \mathbf{W}\ast\mathbf{M}
\end{equation}
where $\ast$ is a convolution operation.

To conduct pixel-wise normalization, for simplicity, let's take $3\times 3$ convolution for an example. $\mathcal{P}=\{(-1,-1),(-1,0),\cdots,(0,1),(1,1)\}$ defines a $3\times 3$ kernel, so
\begin{eqnarray}
    u_{j,k,o} &=& \sum_{\mathbf{p} \in \mathcal{P},i}w_{i,\mathbf{p},o}\cdot m_{(j,k)+\mathbf{p},i} \\
    &=& \sum_{\mathbf{p} \in \mathcal{P},i}w_{o,\mathbf{p},i}\cdot s_{(j,k)+\mathbf{p},i}\cdot c_{(j,k)+\mathbf{p},i} \;.
\end{eqnarray}

Based on the i.i.d. and unit std assumption of each element in $\mathbf{C}$, after modulation and convolution, the std of $u_{j,k,o}$ is
\begin{equation}
    \sigma_{j,k,o} = \sqrt{\sum_{\mathbf{p} \in \mathcal{P},i}w_{o,\mathbf{p},i}^2\cdot s_{(j,k)+\mathbf{p},i}^2} \;.
\end{equation}
It is equivalent to convolute $\mathbf{S}^2$ with ${\mathbf{W}^2}$ and then get square roots element-wisely as
\begin{equation}
    \mathbf{\Sigma} = \sqrt{{\mathbf{W}^2}\ast\mathbf{S}^2}
\end{equation}
where $(\cdot)^2$ and $\sqrt{(\cdot)}$ are all element-wise operations.

Finally, we normalize the output tensor $\mathbf{N}\in\mathbf{R}^{O\times H \times W}$ as
\begin{equation}\label{eqn:modconv}
    \mathbf{N} = \frac{\mathbf{W} \ast (\mathbf{S}\otimes\mathbf{C})}{\sqrt{{\mathbf{W}^2}\ast\mathbf{S}^2}} \;.
\end{equation}


We call \cref{eqn:modconv} as \textit{ModConv}. In practice, as shown in \cref{fig:psm}, several ModConv layers are stacked and then followed by a modulation and $1\times 1$ convolution layer to generate a style-transferred image denoted as $\mathbf{I}_r^{\mathrm{(Mod)}}$, which will then be fused with $\mathbf{I}_r^{\mathrm{(Sam)}}$ by addition to generate a final rendered image $\mathbf{I}_r$.
Due to computation and memory limitations, the size of $\mathbf{I}_r^{\mathrm{(Sam)}}$ is set to $\frac{H}{2}\times\frac{W}{2}$. Before addition, we need to up-sample it to $H\times W$. As shown in \cref{fig:overview}, we get
\begin{equation}
    \mathbf{I}_r = \mathbf{I}_r^{\mathrm{(Mod)}} + \operatorname{UpSample}(\mathbf{I}_r^{\mathrm{(Sam)}}) \;.
\end{equation}

\subsubsection{Attention-based multi-scale style fusion}
\label{sec:amsf}

\begin{figure}[htbp]
  \centering
  \includegraphics[width=\linewidth]{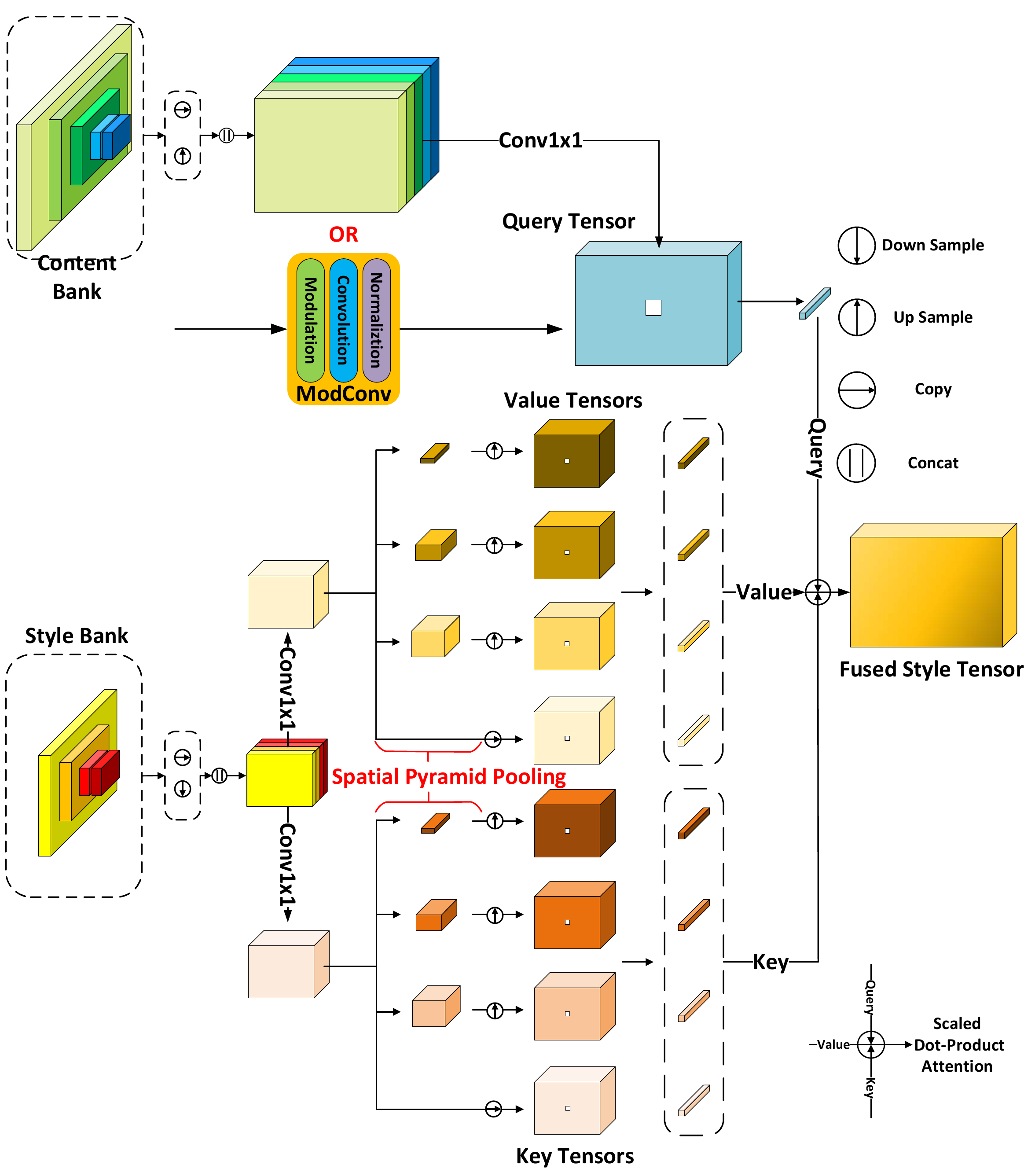}
  \caption{Details of attention-based multi-scale style fusion.}
  \label{fig:amsf}
\end{figure}

Given extracted style feature maps from different stages, generating style tensors for PixyMod is another contribution of this paper.
A natural way to achieve this goal is to up-sample all feature maps in style bank to $H\times W$ and concatenate them together,
then feed this concatenated tensor into a $1\times1$ convolution layer to make its channel number the same as the content tensor in PixyMod.

This strategy works well in cases where the text regions in content and style image align well.
However, when misalignment happens, for example in \cref{fig:baseline_vs_psm}, artifacts happen on the rendered image because respective regions on the style image only contain background pixels. 

To eliminate misalignment-caused artifacts, we design an AttnMuSF module to generate multi-scale style tensors through spatial pyramid pooling (SPP) \cite{SPP_2015_TPAMI} and fuse them with a cross channel attention mechanism spatial-adaptively.

As shown in \cref{fig:amsf}, content tensor, which works as queries in cross channel attention mechanism, is obtained by concatenation and $1\times 1$ convolution, and we denote it as $\mathbf{Q}^{\mathrm{(SF)}}\in\mathbb{R}^{H\times W\times d_f}$. 

Feature maps in style bank are also concatenated first. Due to memory limitation, the shape of this concatenated tensor $\mathbf{S}^{\mathrm{(cat)}}$ is $\frac{H}{8}\times\frac{W}{8}\times480$.
To get key tensors,  a $1\times1$ convolution is then applied to $\mathbf{S}^{\mathrm{(cat)}}$ to generate $\mathbf{K}_0^{\mathrm{(SF)}}\in\mathbb{R}^{\frac{H}{8}\times \frac{W}{8}\times d_f}$.
To capture larger scale style encoding, SPP is applied to $\mathbf{K}_0^{\mathrm{(SF)}}$ and get 
$\mathbf{K}_1^{\mathrm{(SF)}} \in \mathbb{R}^{4\times\frac{4W}{H}\times d_f}$,   
$\mathbf{K}_2^{\mathrm{(SF)}} \in \mathbb{R}^{2\times\frac{2W}{H}\times d_f}$,
$\mathbf{K}_3^{\mathrm{(SF)}} \in \mathbb{R}^{1\times\frac{W}{H}\times d_f}$. 
These 3 tensors are then up-sampled to $\frac{W}{8}\times\frac{H}{8}$. Finally we get 4 key tensors denoted as 
$\mathbf{K}^{\mathrm{(SF)}} = \{\mathbf{K}_i^{\mathrm{(SF)}} | i=0,1,2,3\}$. Similarly, we can get multi-scale value tensors from $\mathbf{S}^{\mathrm{(cat)}}$, denoted as $\mathbf{V}^{\mathrm{(SF)}} = \{\mathbf{V}_i^{\mathrm{(SF)}}\in \mathbb{R}^{\frac{H}{8}\times\frac{W}{8}\times I} | i=0,1,2,3\}$, where $I$ equals to the channel number of content tensor to be modulated in PixyMod.

Given a query vector $\mathbf{q}_{j,k}^{\mathrm{(SF)}}$, its respective key and value vectors will be $\mathbf{K}_{\lfloor \frac{j}{8}\rfloor,\lfloor \frac{k}{8}\rfloor}^{\mathrm{(SF)}} = \{\mathbf{k}_{i,\lfloor \frac{j}{8}\rfloor,\lfloor \frac{k}{8}\rfloor}^{\mathrm{(SF)}} | i=0,1,2,3\}$ and $\mathbf{V}_{\lfloor \frac{j}{8}\rfloor,\lfloor \frac{k}{8}\rfloor}^{\mathrm{(SF)}} = \{\mathbf{v}_{i,\lfloor \frac{j}{8}\rfloor,\lfloor \frac{k}{8}\rfloor}^{\mathrm{(SF)}} | i=0,1,2,3\}$. The fused style vector on this coordinate is as follows:
\begin{equation}
    \mathbf{s}_{j,k} = \left[\operatorname{\mathbf{Softmax}}(\frac{{\mathbf{q}_{j,k}^{\mathrm{(SF)}}}^T\mathbf{K}_{\lfloor \frac{j}{8}\rfloor,\lfloor \frac{k}{8}\rfloor}^{\mathrm{(SF)}}}{\sqrt{d^{{\mathrm{(SF)}}}}})\right]^T\mathbf{V}_{\lfloor \frac{j}{8}\rfloor,\lfloor \frac{k}{8}\rfloor}^{\mathrm{(SF)}} .
\end{equation}

Each modulation operation in PixyMod has its own AttnMuSF module to generate style tensor, whose query tensor is from content bank or previous ModConv layer's output (\cref{fig:psm} and \cref{fig:amsf}).
\subsection{Training strategy}
\subsubsection{Training data generation}
To train APRNet, a dataset containing matched content, style, and ground truth images is required. However, this kind of dataset is difficult to collect. SC-GAN \cite{SCGAN_2020_ICFHR} crops 2 non-overlapping image patches from a real text line image randomly. One serves as the style image, and another as the ground truth image. A skeleton extracted from the ground truth image is used as the content image. This strategy is based on an assumption that the style within a text line image is uniform. Apparently, this assumption does not always hold. 

We propose a new training data generation strategy named \textit{Single Crop} to solve this problem. Different from SC-GAN, given a real text line image, we crop 1 patch from it randomly. This patch serves as the ground truth image and the skeleton extracted from it as the content image. While the style image is generated from ground truth image by the following procedure:
\begin{enumerate}
    \item [1)] Divide the image into $16\times16$ patches;
    \item [2)] For each patch, randomly rotate it by $0^\circ$, $90^\circ$,$180^\circ$ or $270^\circ$;
    \item [3)] For each patch, randomly swap it with one of its 8 neighbor patches.
\end{enumerate}

\begin{figure}[htbp]
  \centering
  \includegraphics[width=0.8\linewidth]{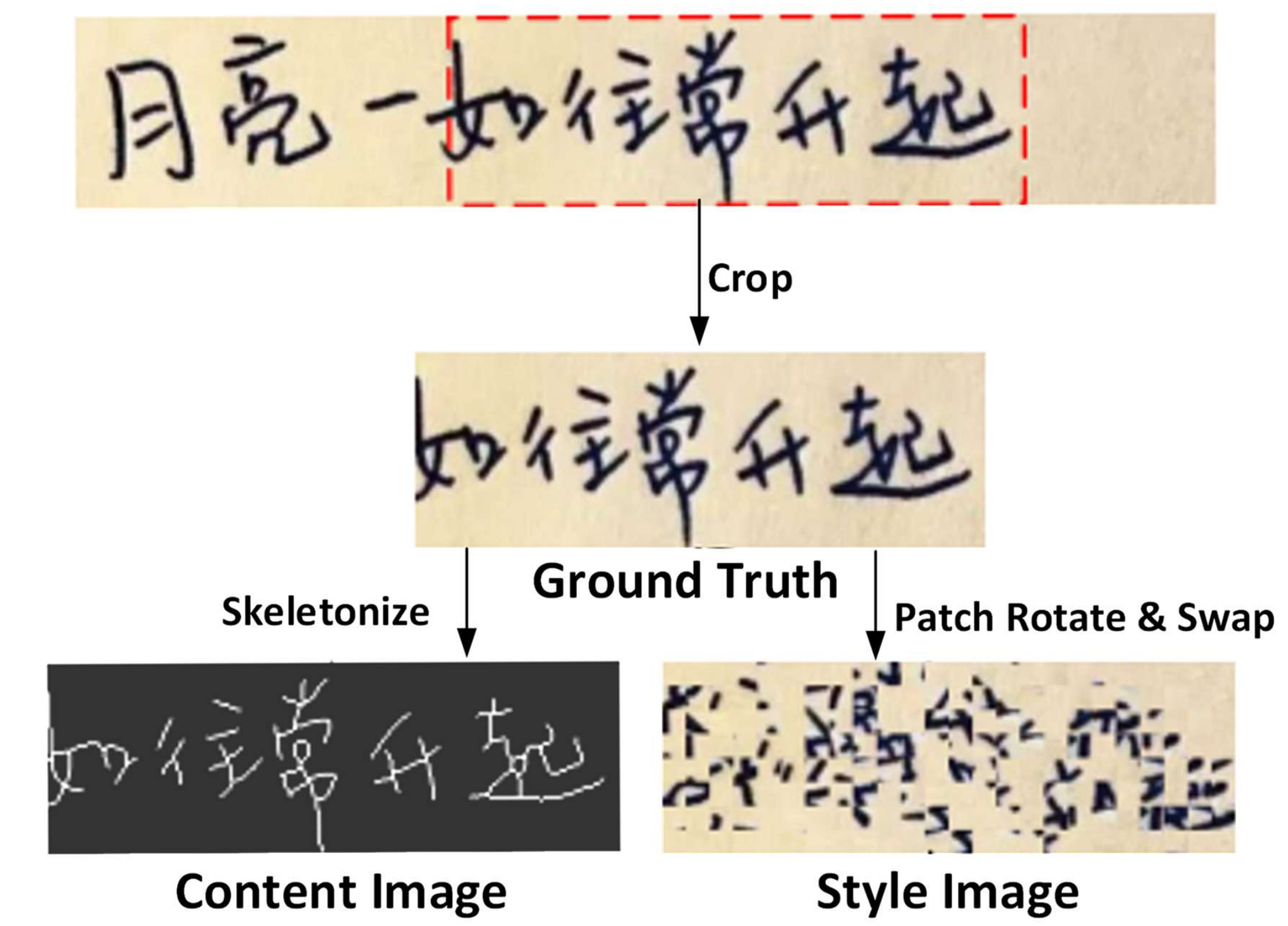}
  \caption{Illustration of \textit{Single Crop} strategy.}
  \label{fig:singlecrop}
\end{figure}

This \textit{Single Crop} method is illustrated in \cref{fig:singlecrop}. With this method, we can create roughly matched content, style, and ground truth images. Patch rotation and shuffling can avoid overfitting and simulate text region misalignment phenomenon in real scenarios. 

\subsubsection{Loss function}
The loss function we used is general, which consists of content loss $\mathcal{L}_c$, perceptual loss $\mathcal{L}_p$ and adversarial loss $\mathcal{L}_a$:
\begin{equation}
    \mathcal{L} = \lambda_c\mathcal{L}_c + \lambda_p\mathcal{L}_p+\lambda_a\mathcal{L}_a
\end{equation}
where $\lambda_{c,p,a}$ are hyper-parameters to control weights of loss components.

We use pixel-wise $L_1$ distance between $\mathbf{I}_r$ and ground truth image $\mathbf{I}_g$ as $\mathcal{L}_c$, 
and mean square error between $\mathbf{I}_r$ and $\mathbf{I}_g$'s 5-stage CNN feature maps extracted by a pretrained VGG-19\cite{Simonyan2014Very} as $\mathcal{L}_p$.
For $\mathcal{L}_a$, we compute adversarial loss for every pixel in discriminator's output\cite{pix2pix_2017_CVPR} to make the synthetic image photo-realistic.

\section{Experiments}
\label{sec:exp}
\subsection{Experimental Setup}
To verify the effectiveness of our design, we evaluate it on a Chinese handwriting text image synthesis task with SCUT-HCCDoc dataset \cite{HCCDoc_2020_PR} and CASIA-OLHWDB dataset \cite{CASIA_2011_ICDAR}.
The SCUT-HCCDoc dataset contains camera-captured document text line images with diverse appearances, backgrounds, and resolutions, which are captured from different application scenarios. 
We use the SCUT-HCCDoc dataset to train our models and serve as the style images in testing. The CASIA-OLHWDB dataset contains online ink trajectories of Chinese text line images. We use the CASIA-OLHWDB2.0-2.2 dataset to create skeleton text line images, which serve as the content images in testing.

Specifically, in the training phase, we randomly split SCUT-HCCDoc dataset into training and validation subsets, which contains 83,926 and 9,325 images, respectively. When constructing a mini-batch, images are first resized to $H = 128$ while keeping aspect ratio, then randomly cropped to $128 \times 384$ patches, which serve as style images. The corresponding content images are obtained using  adaptive threshold binarization and skeletonization methods. 
In the testing phase, we first use the content and style images in the validation set to reconstruct the ground-truth style images. Then we measure their similarities using Peak Signal-to-Noise Ratio (PSNR), Structural Similarity Index Measure (SSIM) \cite{PSNR_SSIM_2010_ICPR}, and LPIPS distance \cite{LPIPS_2018_CVPR} metrics.
We also use CASIA-OLHWDB2.0-2.2 as content images and SCUT-HCCDoc as style images to generate a total of 20,000 synthetic text line images. We measure the similarities between rendered images and corresponding style images using Fréchet inception distance (FID) \cite{FID_2017_NIPS} and Kernel Inception Distance (KID) \cite{KID_2018_ICLR} metrics.


\subsection{Baseline results}

\begin{figure}[htbp]
  \centering
  \includegraphics[width=0.8\linewidth]{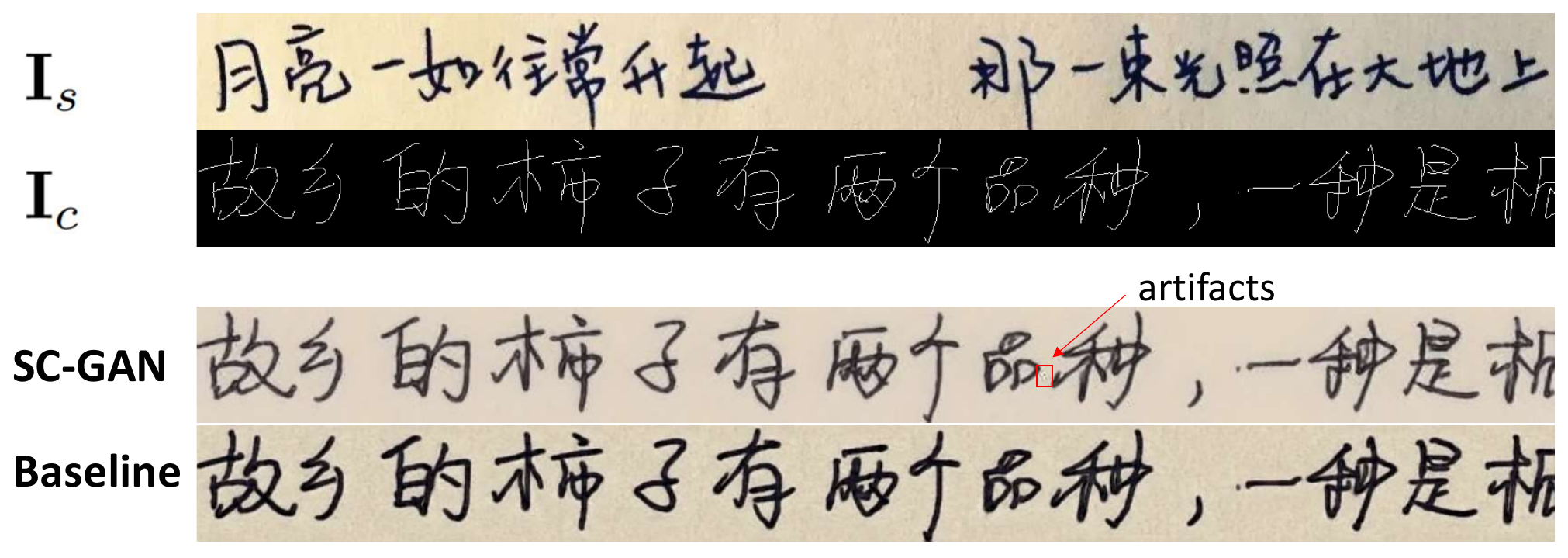}
  \caption{Comparison results between SC-GAN and our StyleGANv2 baseline model. Both models are only able to generate uniform styles across coordinates. There is a ``water droplet'' artifact in the SC-GAN rendered image which is also observed in \cite{StyleGAN2_2020_CVPR}. The StyleGANv2 based model eliminates this artifact and the generated image is more style-consistent with the style reference. }

  \label{fig:sc_gan_vs_baseline}
\end{figure}
We first compare AdaIN-based SC-GAN \cite{SCGAN_2020_ICFHR} model with a StyleGANv2-based \cite{StyleGAN2_2020_CVPR} baseline model. Both models are style vector based. To make the baseline model comparable with APRNet, it also consists of two stages, which generate $\frac{H}{2} \times \frac{W}{2}$ and $H \times W$ images, respectively. In both stages, the style vectors are obtained by directly flattening the last stage's style bank to a vector with global average pooling, followed by a $1\times1$ convolution. 
These vectors are then integrated in the rendering procedure using StyleGANv2's ``demodulation" operation. 
As shown in \cref{fig:sc_gan_vs_baseline}, the lack of spatial information in style encoding makes SC-GAN and baseline model only generate uniform styles across coordinates. ``Water droplet'' artifacts observed in \cite{StyleGAN2_2020_CVPR} also happen in our SC-GAN results. \Cref{fig:sc_gan_vs_baseline} also shows that images rendered by baseline model are more style-consistent with style reference than SC-GAN.  



\subsection{Effect of pixel-wise style modulation}

\begin{figure}[htbp]
  \centering
  \includegraphics[width=0.8\linewidth]{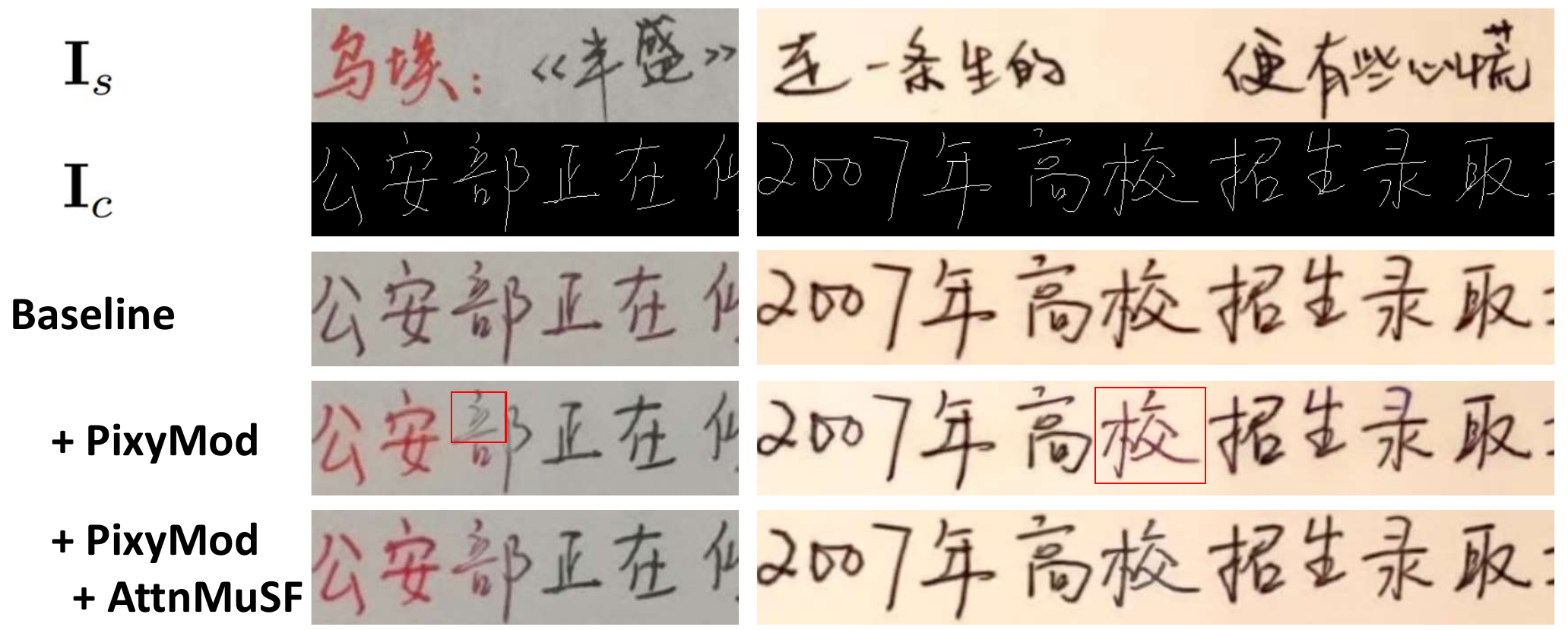}
  \caption{By replacing style vectors with style tensors and integrating style using PixyMod, the spatial-varying styles can be successfully transferred. Equipping PixyMod with AttnMuSF, the rendering quality on misaligned content-style regions is improved.}

  \label{fig:baseline_vs_psm}
\end{figure}

In this section, we replace the style vectors in baseline model with style tensors and transfer spatial-varying color patterns using PixyMod module. This model also consists of 2 rendering stages, and both are based on PixyMod. The style tensors are obtained by up/down-sampling all tensors in style bank to the same spatial size as content tensor and then concatenating them together. The concatenated tensors are fed into $1\times1$ convolutions to match the channel numbers of content tensors. In this way, these style tensors are able to preserve the spatial-varying styles. The comparison results of PixyMod and baseline model are shown in \cref{fig:baseline_vs_psm}. Clearly, compared with baseline, with the help of PixyMod, the spatial-varying styles can be successfully transferred. However, since the style tensors only encode local style information, when the model is trying to render a text region of content image while its corresponding style region only contains background pixels, artifacts will happen, as shown in the red rectangle area of \cref{fig:baseline_vs_psm}. Such misalignment problems will become severe especially when the style image contains wide spaces.


\begin{table*}[htbp]
  \caption{Quantitative comparisons of different methods. $\uparrow$ indicates that higher is better, and $\downarrow$ indicates lower is better. The best results are in bold.}
  \label{tab1}
  \renewcommand\arraystretch{1.0} 
  \footnotesize
  \centering
  \begin{tabular}{l|ccc|cc}
  \hline
  \multirow{2}{*}{Methods} & \multicolumn{3}{c|}{Validation (9,325 pairs)}& \multicolumn{2}{c}{Test (20,000 pairs)} \\
                      & PSNR $\uparrow$  & SSIM $\uparrow$    & LPIPS  $\downarrow$   & FID $\downarrow$      & KID $\downarrow$  ($\times10^{-2}$)   \\ \hline \hline
  SC-GAN (StyleGANv1-based) & 22.03 & 0.826 & 0.130 & 41.17 &  2.54$\pm$0.17   \\
  Baseline (StyleGANv2-based) & 24.54 & 0.862 & 0.109 & 40.71 &  2.34$\pm$0.13   \\
  \quad + PixyMod        & 27.14 & 0.886 & 0.083 & 39.41 &  2.48$\pm$0.13  \\
  \quad + PixyMod + AttnMuSF & 26.60 & 0.879 & 0.090 & 33.19 &  1.80$\pm$0.11   \\
  \quad + PixyMod + AttnPixamp & \textbf{27.41} & 0.891 & \textbf{0.078} & 33.18 & 1.94$\pm$0.12  \\
  \quad + PixyMod + AttnMuSF + AttnPixamp (APRNet) & 27.09 & \textbf{0.892} & 0.084  & \textbf{32.96} &   \textbf{1.74$\pm$0.10}  \\
  \hline
  \end{tabular} 
\end{table*}

\subsection{Effect of attention-based multi-scale style fusion}

In order to eliminate the misalignment-caused artifacts, we use the proposed AttnMuSF module to fuse multi-scale style encodings into style tensor. Ideally, at misaligned regions, AttnMuSF will pay more attention to larger scale style encoding. 

As shown in \cref{fig:baseline_vs_psm}, the rendering quality on the misaligned regions has been improved greatly.
We also visualize the attention map of APRNet's AttnMuSF module on different encoding scales in \cref{fig:effect_vs_attnmusf}. It's shown that if a text region in content is aligned to background one in style, AttnMuSF will assign more weights on $1\times\frac{W}{H}$ encoding, otherwise local style encodings will be attended, which verifies that AttnMuSF works just as expected.




\subsection{Effect of attention-based pixel sampling}

\begin{figure}[htbp]
  \centering
  \includegraphics[width=0.8\linewidth]{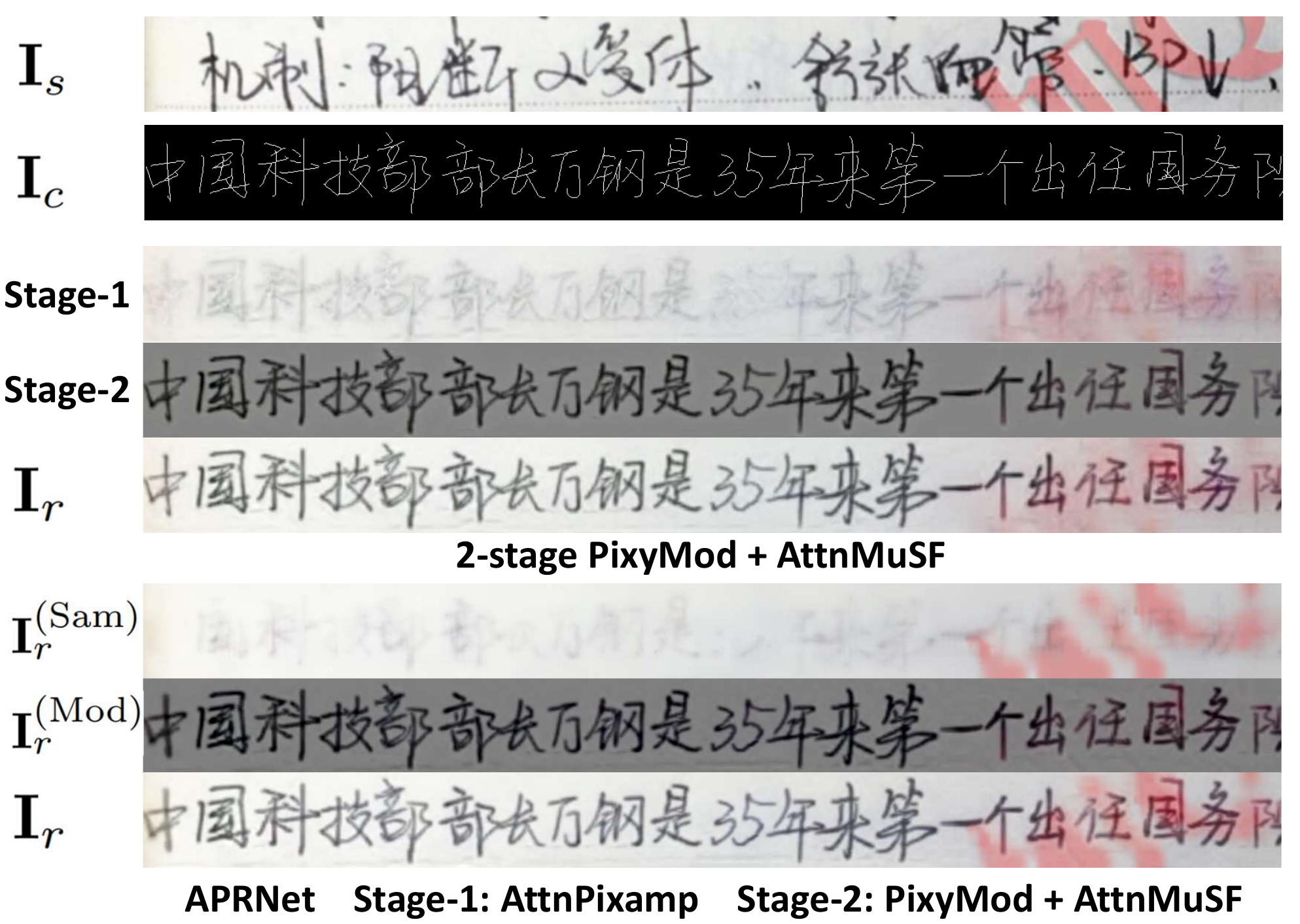}
  \caption{Effect of AttnPixamp module. 2-stage ``PixyMod+AttnMuSF'' generates blurry background. Replacing the 1st stage with AttnPixamp, the background quality of rendered image can be improved. }
  \label{fig:effect_vs_attnpixamp}
\end{figure}

\begin{figure}[htbp]
  \centering
  \includegraphics[width=0.8\linewidth]{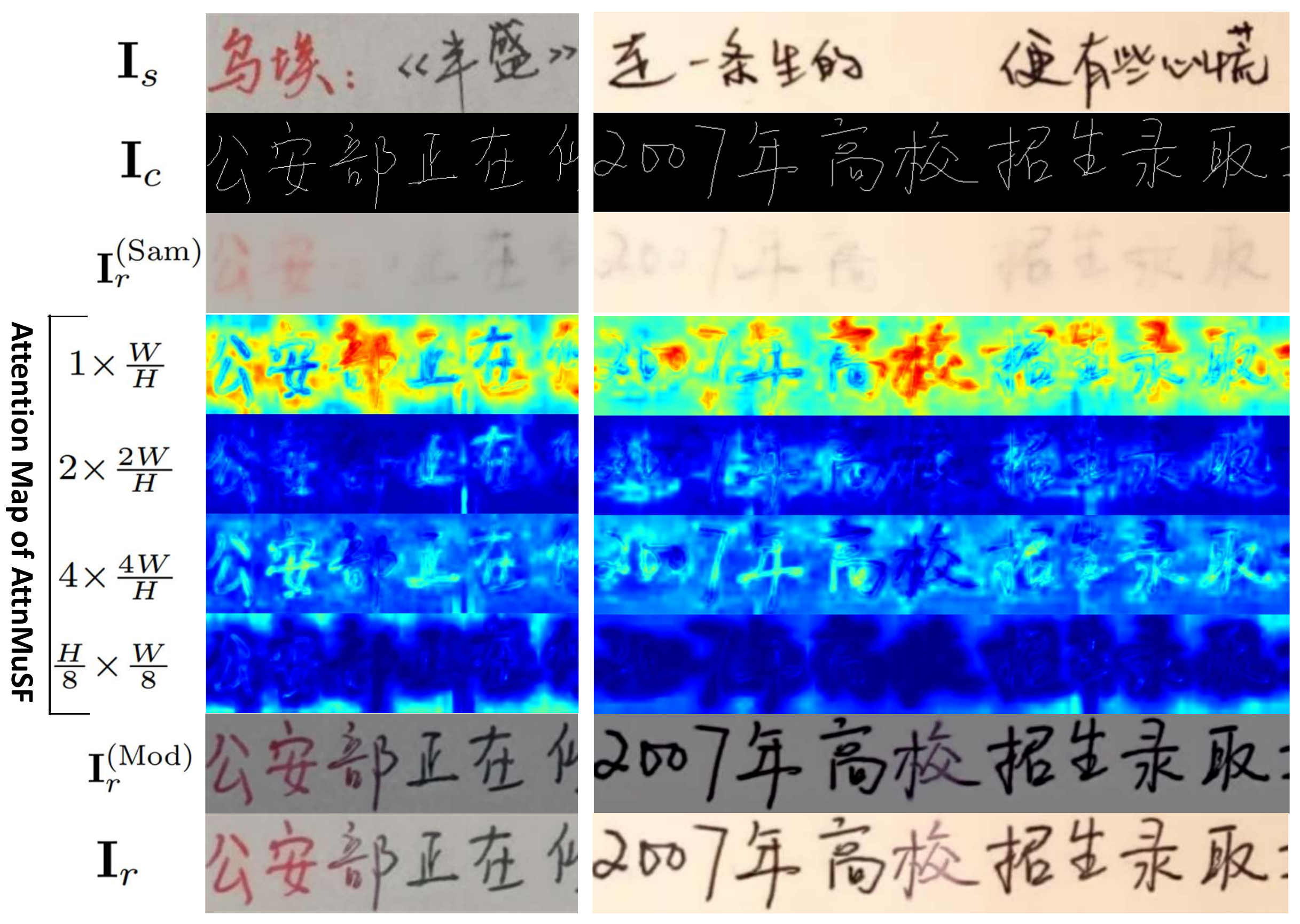}
  \caption{ APRNet's 1st stage AttnPixamp ($\mathbf{I}_r^{\text{(Sam)}}$) and 2nd stage ``PixyMod+AttnMuSF'' ($\mathbf{I}_r^{\text{(Mod)}}$) outputs, and the attention map visualizations of AttnMuSF in the 2nd stage. The AttnPixamp stage mainly focuses on background and the ``PixyMod+AttnMuSF'' stage focuses on content foreground. When a text region is aligned to background in style, AttnMuSF will pay more attention to the $1 \times \frac{W}{H} $ encodings.}
  \label{fig:effect_vs_attnmusf}
\end{figure}

However, a 2-stage ``PixyMod+AttnMuSF'' model still cannot handle complex background rendering well. As shown in \cref{fig:effect_vs_attnpixamp},
when the background styles become complex, both stages generate images with a blurry background.  Therefore we replace the 1st stage with an AttnPixamp module. For each coordinate, we sample 25 pixels from its $17\times17$ neighborhood grid ($k=5,m=4$) uniformly, and aggregate these pixels by style-content cross attention to get render result.   \cref{fig:effect_vs_attnpixamp} shows that AttnPixamp can improve background rendering quality greatly. Intermediate results in \cref{fig:effect_vs_attnpixamp}
and \cref{fig:effect_vs_attnmusf} also show that with AttnPixamp focusing on background rendering, PixyMod can pay more attention to the foreground. This disentanglement effect improves final rendering quality and makes the output more photo-realistic.

\subsection{Qualitative comparison}

\begin{figure*}[htbp]
  \centering
  \includegraphics[width=0.8\linewidth]{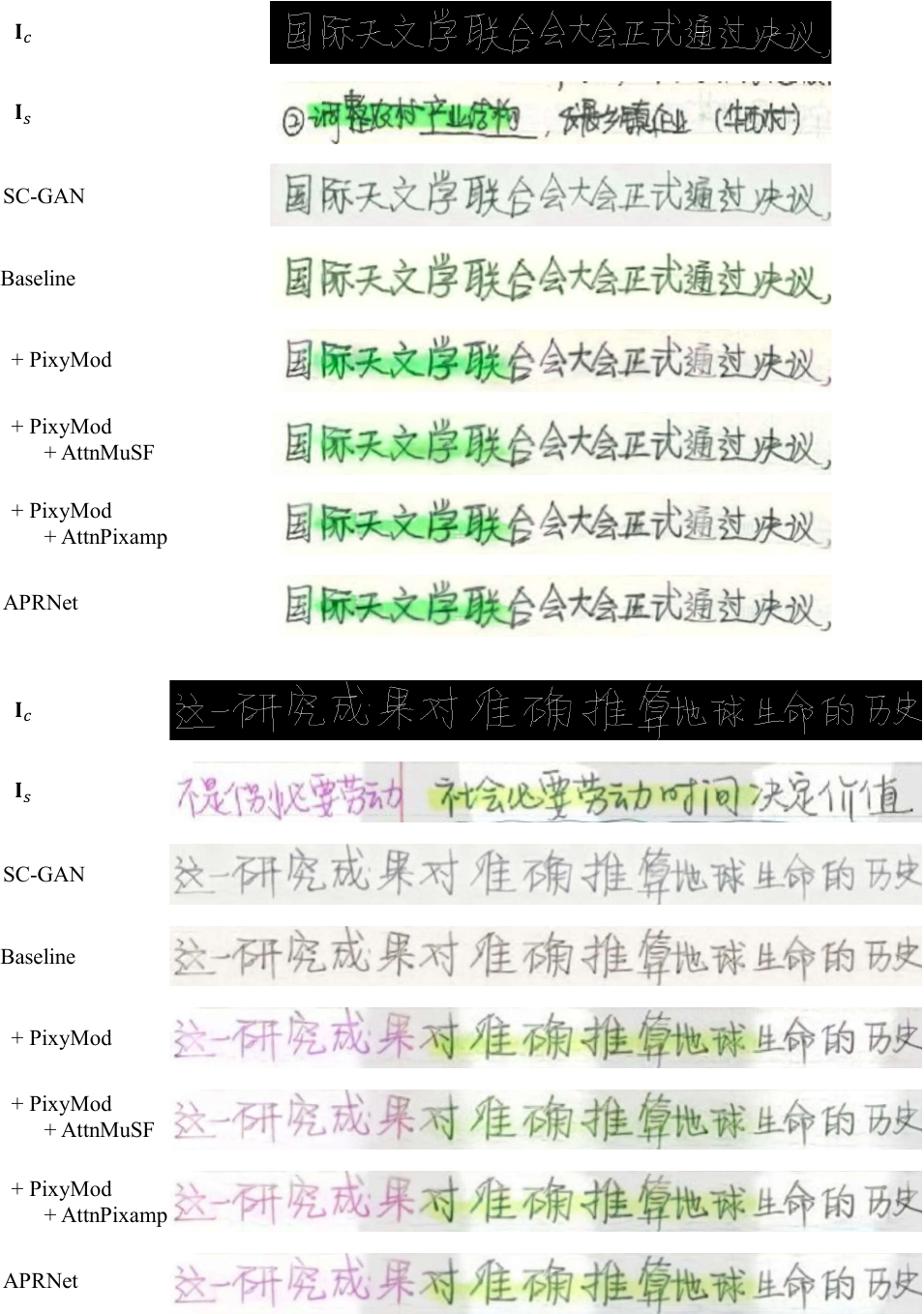}
  \caption{Additional results with comparison to other models.}
  \label{fig:example}
\end{figure*}

\begin{figure*}[htbp]
  \centering
  \includegraphics[width=0.8\linewidth]{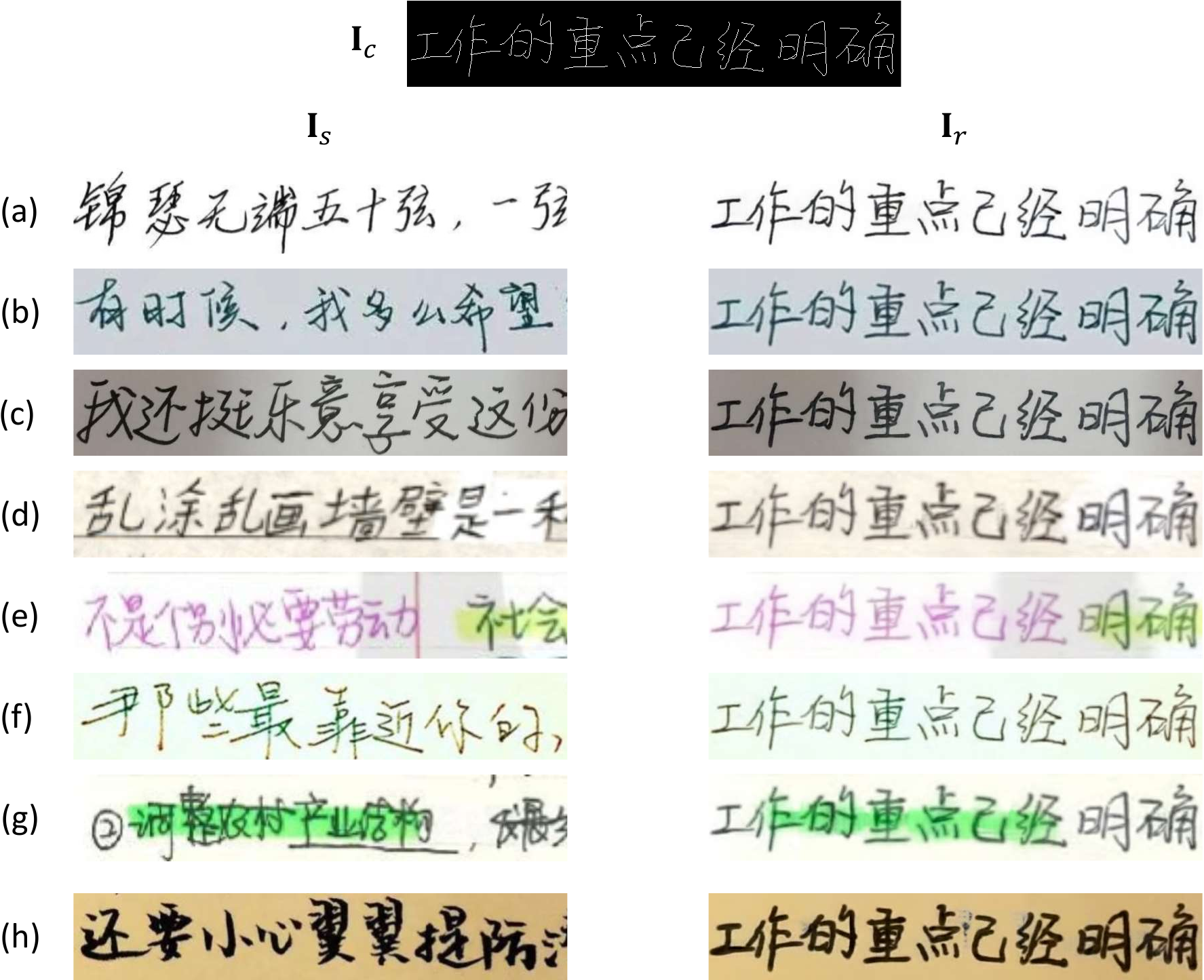}
  \caption{Additional results of APRNet using the same content image and different style images. APRNet is able to imitate reference images with (a-b) simple and uniform foreground and backgrounds, (c-f) spatial-varying styles, and (g) complex backgrounds. However, there can be a ``blob-like'' artifact cause by AttnPixamp, when all candidates pixels from style images of AttnPixamp lie in foregrounds.}
  \label{fig:same_content_diff_style}
\end{figure*}

\cref{fig:example} shows the generation results of the APRNet and other models when inputting complex style images. The APRNet significantly improves the generation quality compared with the SC-GAN and the baseline model, and performs better than other models in terms of the details. \cref{fig:same_content_diff_style} presents the synthesis results of APRNet using the same content image and different style images. Clearly, the APRNet is able to imitate style images with simple/uniform and complex/spatial-varying styles. The synthetic results in \cref{fig:same_content_diff_style} demonstrate that the stroke width of handwritten text can also be transferred. However, there can be a ``blob-like'' artifact when the style image contains thick and dense text strokes. After careful analysis, we find that it is caused by the AttnPixamp module. Because we are using uniform-sampled pixels as candidates for each pixel in content bank, there are chances when the pixel in content bank is background while all sampled candidates in the style image lie in foregrounds. This issue can be solved if the candidates are sampled adaptively. We leave this as our future works.

\subsection{Quantitative comparison}
Finally, we compare these models using quantitative measurements. Results are listed in \cref{tab1}. Besides ``PixyMod'', ``PixyMod+AttnMuSF'' and APRNet, we also list the result of ``PixyMod+AttnPixamp'' in which we remove the AttnMuSF in APRNet.
According to \cref{tab1}, the PSNR, SSIM, and LPIPS scores are greatly improved using PixyMod. It indicates that using PixyMod, models are able to imitate the spatial-varying styles in reference style images much better, which achieve lower reconstruction differences. The FID and KID scores on the testing set are improved a lot using AttnMuSF and AttnPixamp. It shows that using these two methods, the generated images are more photo-realistic and visually similar to corresponding style reference. Combining PixyMod, AttnMuSF and AttnPixamp, our proposed APRNet achieves much better results than SC-GAN and baseline models.

\section{Limitations}
\label{sec:lim}
Our APRNet still has the following limitations:
\begin{itemize}
    \item [1)] Sharp edges in background are not rendered very well;
    
    \item [2)] Memory requirement of APRNet is pretty large;
    
    \item [3)] \textit{Single Crop} strategy releases us from collecting matched tuple images for training, but it relies on high quality binarization/skeletonization techniques to achieve satisfactory results;
    
    \item [4)] APRNet can only transfer color patterns currently. It can not imitate style image's typography/writing styles, spatial transformations/deformations, etc.
\end{itemize}

\section{Conclusion}
\label{sec:con}
We propose an APRNet equipped with AttnPixamp, PixyMod and AttnMuSF modules to transfer spatial-varying color patterns from style text image to content image and a \textit{Single Crop} strategy to train this model.
Experimental results on Chinese handwriting text image synthesis show that our design can improve the quality of synthetic text images and make them more photo-realistic.

{
\bibliographystyle{IEEEtran}
\bibliography{ref/refs_handwritten_generation, ref/refs_style_transfer, ref/refs_printed_generation, ref/refs, ref/refs_metric}
}

\appendices

\section{Additional implementation details}
\begin{figure*}[htbp]
  \centering
  \includegraphics[width=0.8\linewidth]{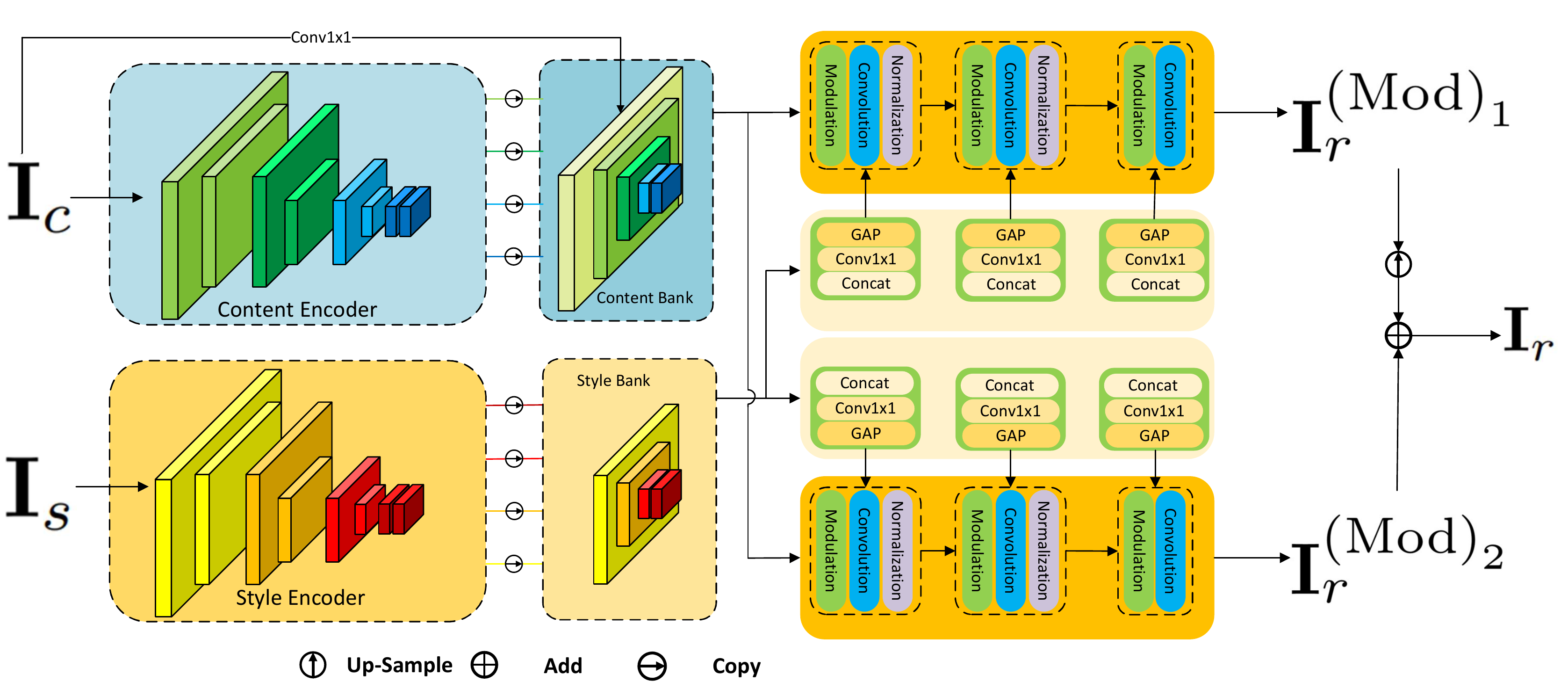}
  \caption{StyleGANv2-based baseline system. This system consists of 2 rendering stages, which will generate $\frac{H}{2}\times\frac{W}{2}$ and $H\times W$ images respectively. The style \textbf{vectors} used for modulation and normalization are generated by 1) down/up-sampling all feature maps in style bank to same size, 2) concatenating them to form a raw style tensor, 3) feeding this tensor to a $1\times 1$ conv layer to fit the channel number of content tensors, and 4) generating style vectors through global average pooling (\textbf{GAP}). }
  \label{fig:baseline}
\end{figure*}

\begin{figure*}[htbp]
  \centering
  \includegraphics[width=0.8\linewidth]{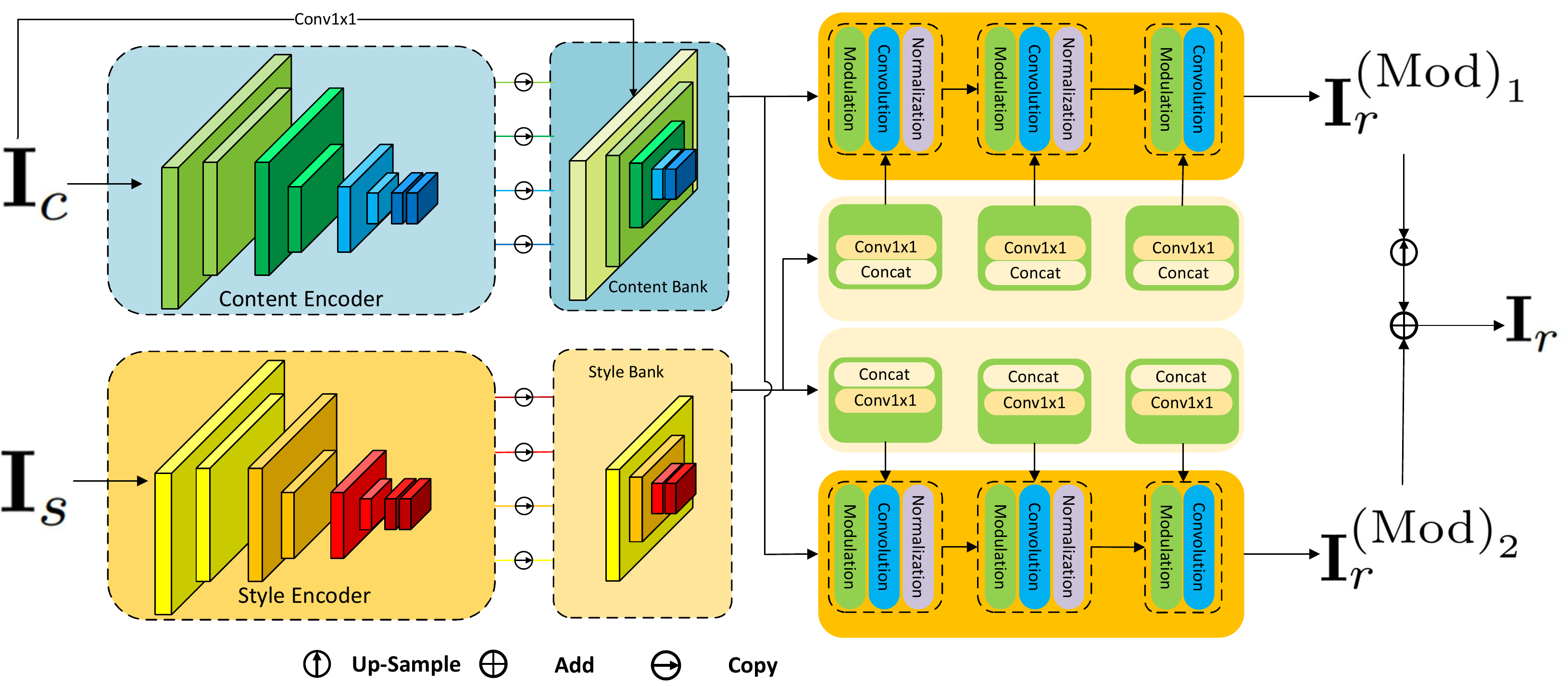}
  \caption{``Baseline+PixyMod'' system. Compared with \cref{fig:baseline}, this system replaces style vectors with style tensors. It is implemented by abandoning the GAP operations.}
  \label{fig:baseline_pixymod}
\end{figure*}

\begin{figure*}[htbp]
  \centering
  \includegraphics[width=0.8\linewidth]{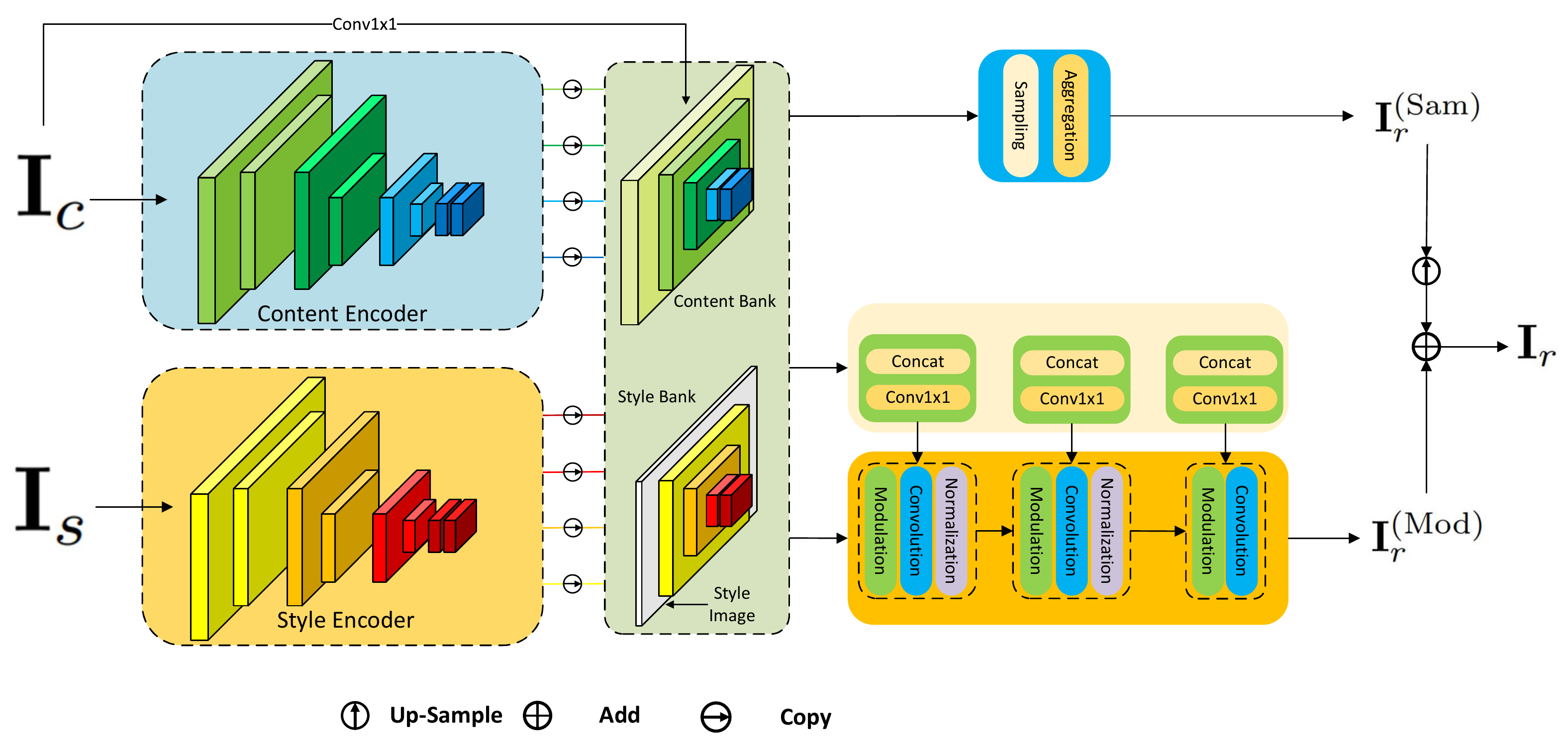}
  \caption{``Baseline+PixyMod+AttnPixamp'' system. Compared with \cref{fig:baseline_pixymod}, this system replaces first stage with AttnPixamp Module.}
  \label{fig:baseline_attnpixamp_pixymod}
\end{figure*}

\begin{figure*}[htbp]
  \centering
  \includegraphics[width=0.8\linewidth]{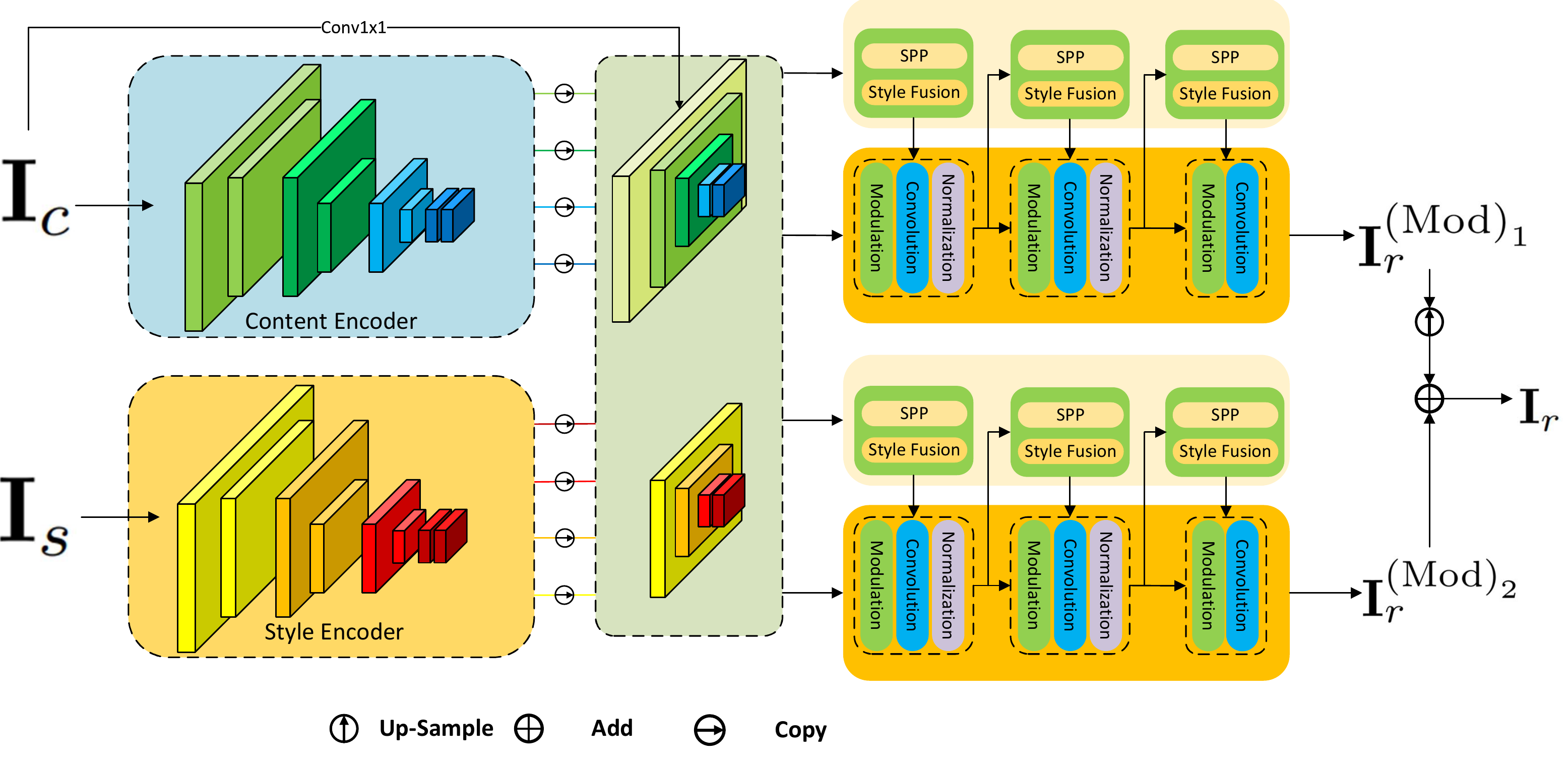}
  \caption{``Baseline+PixyMod+AttnMuSF'' system. Compared with \cref{fig:baseline_pixymod}, this system generates style vectors via AttnMuSF module.}
  \label{fig:baseline_attnmusf_pixymod}
\end{figure*}

We add more details of all systems we compared including StyleGANv2-based baseline system (\cref{fig:baseline}), ``Baseline+PixyMod'' system (\cref{fig:baseline_pixymod}), ``Baseline+PixyMod+AttnPixamp'' system (\cref{fig:baseline_attnpixamp_pixymod}) and ``Baseline+PixyMod+AttnMuSF'' system (\cref{fig:baseline_attnmusf_pixymod}).



\end{document}